\documentclass[sigconf]{acmart}

\usepackage{algpseudocode}
\usepackage{algorithm}
\usepackage{graphicx}
\usepackage{bm}
\usepackage{multirow}
\usepackage{bbding}
\usepackage{pifont}
\usepackage{url}
\usepackage{amsmath}

\usepackage{xcolor}         
\usepackage{multirow}
\usepackage{booktabs}
\usepackage{amssymb}
\usepackage[table]{xcolor}
\usepackage{balance}

\AtBeginDocument{%
  }

\begin{document}

\title{MixedGaussianAvatar: Realistically and Geometrically Accurate Head Avatar via Mixed 2D-3D Gaussians}

\author{Peng Chen}
\affiliation{%
  \institution{University of the Chinese Academy of Sciences}
  \city{Beijing}
  \state{Beijing}
  \country{China}
}
\email{chenpeng23@mails.ucas.ac.cn}

\author{Xiaobao Wei}
\affiliation{%
  \institution{University of the Chinese Academy of Sciences}
  \city{Beijing}
  \state{Beijing}
  \country{China}}
\email{weixiaobao23@mails.ucas.ac.cn}

\author{Qingpo Wuwu}
\affiliation{%
  \institution{Peking University}
  \city{Beijing}
  \state{Beijing}
  \country{China}}
\email{2401112105@stu.pku.edu.cn}

\author{Xinyi Wang}
\affiliation{%
  \institution{Nankai University}
  \city{Tianjin}
  \state{Tianjin}
  \country{China}}
\email{wangxinyi.nemu@mail.nankai.edu.cn}

\author{Xingyu Xiao}
\affiliation{%
  \institution{Tsinghua University}
  \city{Beijing}
  \state{Beijing}
  \country{China}}
\email{xxy23@mails.tsinghua.edu.cn}

\author{Ming Lu}
\affiliation{%
  \institution{Intel Labs China}
  \city{Beijing}
  \state{Beijing}
  \country{China}}
\email{lu199192@gmail.com}


\begin{abstract}
Reconstructing high-fidelity 3D head avatars is crucial in various applications such as virtual reality. The pioneering methods reconstruct realistic head avatars with Neural Radiance Fields (NeRF), which have been limited by training and rendering speed. Recent methods based on 3D Gaussian Splatting (3DGS) significantly improve the efficiency of training and rendering. However, the surface inconsistency of 3DGS results in subpar geometric accuracy; later, 2DGS uses 2D surfels to enhance geometric accuracy at the expense of rendering fidelity. To leverage the benefits of both 2DGS and 3DGS, we propose a novel method named MixedGaussianAvatar for realistically and geometrically accurate head avatar reconstruction. Our main idea is to utilize 2D Gaussians to reconstruct the surface of the 3D head, ensuring geometric accuracy. We attach the 2D Gaussians to the triangular mesh of the FLAME model and connect additional 3D Gaussians to those 2D Gaussians where the rendering quality of 2DGS is inadequate, creating a mixed 2D-3D Gaussian representation. These 2D-3D Gaussians can then be animated using FLAME parameters. We further introduce a progressive training strategy that first trains the 2D Gaussians and then fine-tunes the mixed 2D-3D Gaussians. We use a unified mixed Gaussian representation to integrate the two modalities of 2D image and 3D mesh. Furthermore, the comprehensive experiments demonstrate the superiority of MixedGaussianAvatar. The code will be released.
\end{abstract}




\maketitle
\section{Introduction}
Reconstructing high-quality 3D head avatars from multiple 2D images captured from different viewpoints is essential for various applications~\cite{wohlgenannt2020virtual,chollet2009multimodal, bai2024bring, wei2024gazegaussian, chen2025diffusiontalker, shao2024splattingavatar, wei2025graphavatar}. The key high-fidelity aspects include the geometric and realistic accuracy of 3D head avatars and rendered images. With the advancement of deep learning, techniques based on Neural Radiance Fields (NeRF)~\cite{mildenhall2021nerf} and 3D Gaussian Splatting (3DGS)~\cite{kerbl20233d} have gained popularity, providing significant benefits for creating high-fidelity 3D head avatars.
\begin{figure*}[t]
    \centering
    \includegraphics[width=0.9\textwidth]{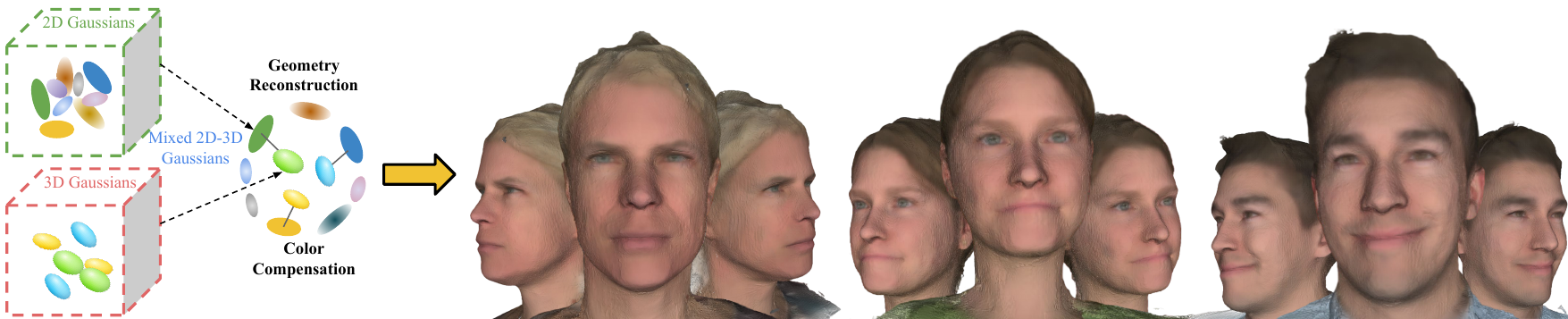}
    \caption{MixedGaussianAvatar uses a mixed 2D-3D Gaussian Splatting method to reconstruct a realistically and geometrically accurate 3D head avatar mesh.}
    \label{fig:teaser}
    \vspace{-2mm}
\end{figure*}
NeRF introduces the idea of radiance fields, utilizing neural networks to store 3D information, which allows for reconstructing high-quality static scenes. When it comes to 3D head avatar reconstruction, the proposed methods primarily focus on reconstructing dynamic head avatars~\cite{Gao2022nerfblendshape, Zielonka2022InstantVH}. For example, Neural Head Avatars (NHA)~\cite{grassal2022neural} process RGB video inputs that display various expressions and perspectives. NHA involves estimating and refining the low-dimensional shape, expression, and pose parameters of the FLAME head model from an RGB input frame, which allows for creating a neural head avatar that can be animated and rendered from novel viewpoints. 
However, the head avatars generated using NeRF-based methods have lower accuracy due to the implicit representation characteristics of NeRF, and the training and rendering processes are slow.

Recently, 3DGS effectively models a static scene using 3D Gaussians and rendered it through a differentiable rasterizer, greatly speeding up the 3D reconstruction process from multi-view RGB images. Because of 3DGS’s explicit representation and efficient rasterization properties, it achieves high rendering quality and fast training and inference speeds~\cite{chen2024monogaussianavatar, wang2025plgs, hu2024gauhuman, hu2024gaussianavatar}. 
In 3D head avatar reconstruction, most methods attach 3D Gaussians to the FLAME triangular mesh to create dynamic facial expressions~\cite{yu2024gaussiantalker}.
For example, GaussianAvatars~\cite{qian2024gaussianavatars} employ a rigging approach to align the local positions, rotations, and scales of 3D Gaussians with the global parameters. FlashAvatar~\cite{xiang2024flashavatar} and Gaussian Head Avatar ~\cite{xu2024gaussian} utilize neural networks to predict 3D Gaussian parameters. The advantage of these methods lies in their ability to dynamically represent 3D head avatars while producing high-quality images. However, due to the inherent multi-view inconsistency of 3DGS, it cannot accurately reconstruct geometric surfaces.

2D Gaussian Splatting~\cite{huang20242d} replaces 3D Gaussians with 2D Gaussians and employs a ray-splat intersection method for the splatting process, ensuring consistency across multiple views. The advantage lies in the high quality of reconstructed geometric surfaces, but it compromises the quality of rendered images. To summarize 3DGS and 2DGS, 1) 3DGS has achieved realistic results in novel view synthesis, but it struggles to capture accurate geometric structures. 2) 2DGS uses 2D surfels to accurately reconstruct geometric surfaces, but it cannot render realistic images as effectively as 3DGS. 

To unify the two modalities of 2D image rendering and 3D mesh reconstruction, we harness the advantages of both 3DGS and 2DGS and introduce an innovative method called MixedGaussianAvatar. 
This method uses a mixed 2D-3D Gaussians representation, combining the color rendering advantages of 3DGS with the geometric reconstruction strengths of 2DGS to achieve a realistically and geometrically accurate 3D head avatar reconstruction. 
As shown in Fig.~\ref{fig:teaser}, our primary goal is to use 2DGS to preserve the geometric accuracy of the 3D head surface. We attach 2D Gaussians to the triangular mesh of the FLAME model and then connect additional 3D Gaussians to the corresponding 2D Gaussians in areas where the rendering quality is insufficient. The mixed 2D-3D Gaussians can be driven using FLAME parameters to form a dynamic 3D representation. To train the mixed 2D-3D Gaussians, we further propose a progressive training strategy that first trains the 2D Gaussians and then finetunes the mixed 2D-3D Gaussians. 
Our method unifies image rendering and mesh reconstruction through a mixed Gaussian representation, with contributions summarized as follows:

\begin{itemize}
\item We introduce MixedGaussianAvatar, an innovative approach that combines the rendering benefits of 3DGS with the reconstruction strengths of 2DGS, enabling the realistically and geometrically accurate reconstruction of 3D head avatars.
\item We present a progressive training strategy to train mixed 2D-3D Gaussians for dynamic 3D head avatars.
\item We conduct extensive experiments that show MixedGaussianAvatar achieves state-of-the-art color rendering and geometric reconstruction results.
\end{itemize}

\section{Related Work}

\begin{figure*}[h]
    \centering
    \includegraphics[width=0.9\textwidth]{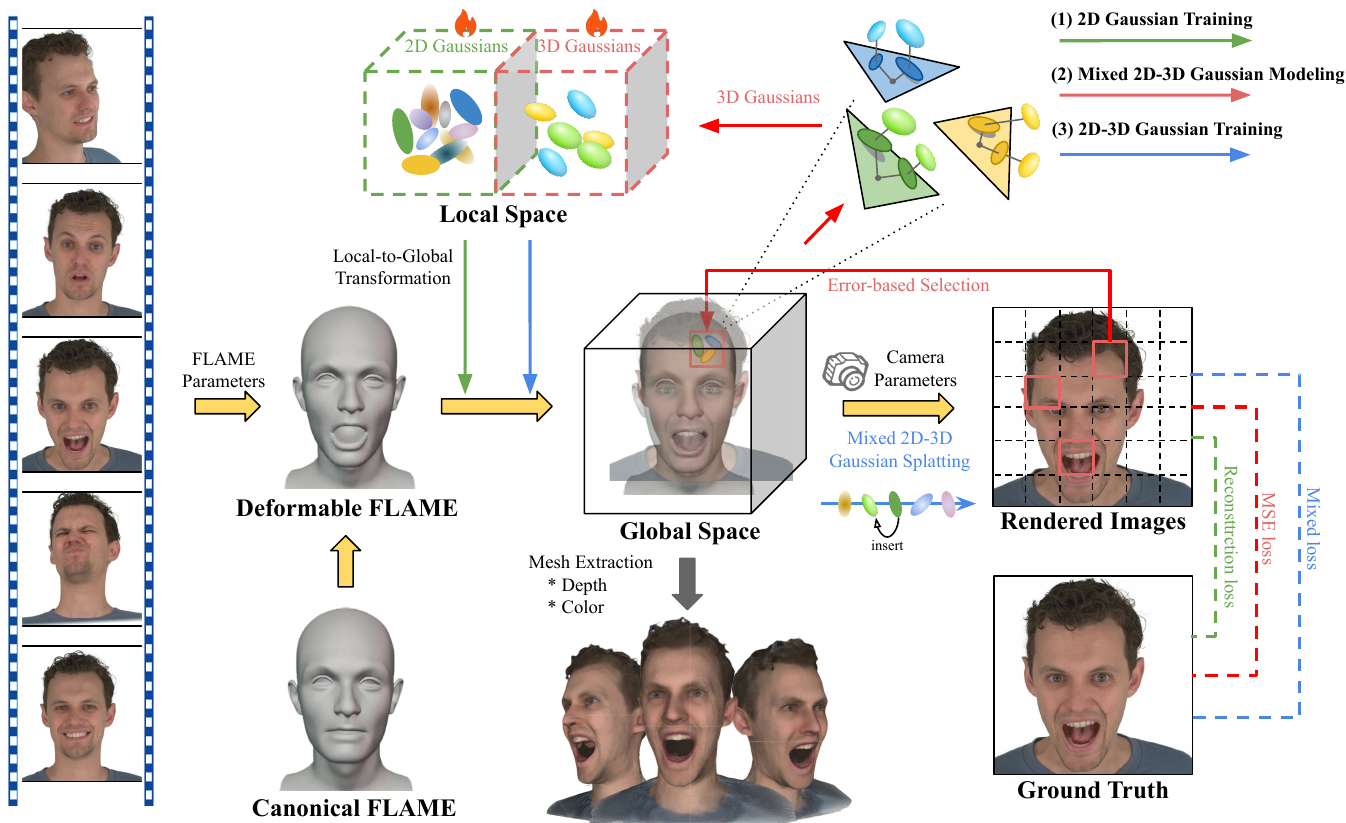}
    \caption{\textbf{Pipeline of MixedGaussianAvatar.}
    We propose a differentiable mixed 2D-3D Gaussian Splatting method for reconstructing realistically and geometrically accurate head avatars from multi-view 2D images. This approach uses 2D Gaussians for geometric precision and 3D Gaussians to correct color errors, resulting in high-quality 3D head mesh and realistic mesh textures or RGB images. The method is integrated with the FLAME model, enabling dynamic effects and animation driven by parameters through local-to-global transformation and the progressive training strategy.
    }
    \label{fig:pipeline}
    \vspace{-2mm}
\end{figure*}

\paragraph{Dynamic Neural Fields}
The representation of dynamic scenes has attracted considerable attention. Previous methods, like NeRF~\cite{mildenhall2021nerf} and its variants~\cite{muller2022instant, chen2022tensorf, wei2024nto3d, wei2024emd}, focus on modeling static scenes, storing them in learnable neural networks to achieve competitive novel view synthesis quality. Recently, 3D Gaussian Splatting (3DGS)~\cite{kerbl20233d} and its variants~\cite{lu2024scaffold} have introduced a much faster training and rendering pipeline. The methods mentioned above are only effective for modeling static scenes and have difficulty handling dynamic scenes. Based on NeRF and 3DGS, several approaches~\cite{shao2023tensor4d, wu20244d, huang2024textit} introduce 4D coordinates to effectively represent dynamic scenes. Humans are very sensitive to facial details, making it challenging for general methods to create high-fidelity head avatars. Recently, numerous efforts~\cite{grassal2022neural, hong2022headnerf, chen2023diffusiontalker} have been made to develop specialized techniques for representing 3D head avatars. AvatarMAV~\cite{xu2023avatarmav} and INSTA~\cite{zielonka2023instant}, both based on voxel representations, enable fast avatar rendering. PointAvatar~\cite{zheng2023pointavatar} is the first to employ a deformable point-based representation for creating high-fidelity avatars. Inspired by 3DGS and PointAvatar, methods like GaussianAvatars~\cite{qian2024gaussianavatars}, Gaussian Head Avatar~\cite{xu2024gaussian}, and FlashAvatar~\cite{xiang2024flashavatar} attach 3D Gaussians to dynamic meshes or UV maps, achieving success in real-time 3D head avatar animation. Previous methods fail to consider geometric reconstruction, which makes them less appropriate for industrial applications. 
In contrast, MixedGaussianAvatar integrates the high-quality appearance of 3DGS with the consistent geometry of 2DGS.

\paragraph{Head Avatar Reconstruction.}
Compared to neural rendering, which primarily focuses on view synthesis, neural reconstruction emphasizes explicit surface extraction and textured mesh reconstruction~\cite{aneja2025scaffoldavatar, qiu2025anigs, lee2024surfhead, moon2025geoavatar, zielonka2023drivable, zhang2024rodinhd, li2024animatable, liu2024humangaussian, qian20243dgs, metzer2023latent}. 
NeuS~\cite{wang2021neus} is the first to propose a bias-free volume rendering method based on neural fields that leverages a signed distance function (SDF) to achieve high-quality surface reconstruction.
UNISURF~\cite{oechsle2021unisurf} presents a unified framework that combines neural implicit surfaces and radiance fields to enable accurate 3D surface reconstruction from multi-view images without requiring object masks.
VolSDF~\cite{yariv2021volume} significantly improves geometry approximation and disentangles shape and appearance compared to previous neural rendering methods.
With the development of 3D Gaussian Splatting, SuGaR~\cite{guedon2024sugar} introduces a method for fast and precise 3D mesh reconstruction and high-quality mesh rendering by aligning 3D Gaussian splats with scene surfaces, enabling efficient and detailed mesh extraction within minutes.
2D Gaussian Splatting (2DGS)~\cite{huang20242d} proposes a novel approach for geometrically accurate radiance field reconstruction by leveraging 2D Gaussian primitives, which significantly improves surface modeling and view-consistent rendering compared to 3D Gaussian splatting methods.
These works focus on the general static object surface reconstruction. 
Within the head avatar reconstruction, NHA~\cite{grassal2022neural} presents a novel approach to reconstructing full human head geometry and photorealistic textures from a short monocular RGB video.
Ming et al.~\cite{ming2024high} introduces a technique for reconstructing personalized, animation-ready mesh blendshapes from single or sparse multi-view videos using neural inverse rendering.

To our knowledge, we are the first to utilize mixed Gaussians to explicitly represent head avatar meshes. Compared to 3DGS-based methods like GaussianAvatars, which exhibit poorer surface reconstruction, our approach mixes the advantages of 2DGS and 3DGS to achieve an optimal balance between rendering and reconstruction quality.
\section{Method}
\subsection{Preliminaries}
3DGS represents 3D scenes using Gaussian primitives, where each primitive is described by a 3D position \(\mathbf{p_k}\) and a 3D covariance matrix \(\Sigma\). The Gaussian function is defined as:
\begin{equation}
G(\mathbf{p}) = \exp\left(-\frac{1}{2} (\mathbf{p} - \mathbf{p_k})^\top \Sigma^{-1} (\mathbf{p} - \mathbf{p_k})\right)
\end{equation}
where the covariance matrix \(\Sigma\) is decomposed into a rotation matrix \(R\) and a scaling matrix \(S\), given by \(\Sigma = R S S^\top R^\top\). During the rendering process, these 3D Gaussian primitives are projected onto the image plane using a world-to-camera transformation matrix \(W\) and a local affine transformation matrix \(J\). This projection yields projected 2D Gaussian primitives with a covariance matrix \(\Sigma^{2D}\), derived from \(\Sigma\) after omitting its third row and column. Then we can obtain $G^{3D}$ with \(\Sigma^{2D}\).
Finally, to render the scene, 3DGS uses volumetric alpha blending, where the contributions of each Gaussian primitives are combined sequentially from front to back to determine the color at a pixel \(\mathbf{x}\):
\begin{equation}
\mathbf{c}(\mathbf{x}) = \sum_{k=1}^{K} \mathbf{c_k} \alpha_k G^{3D}_k(\mathbf{x}) \prod_{j=1}^{k-1} (1 - \alpha_j G^{3D}_j(\mathbf{x}))
\label{eq:rendering}
\end{equation}
where \(\mathbf{c_k}\) represents the color, and \(\alpha_k\) is the alpha value of each Gaussian primitive.

2DGS removes the third dimension of scaling from the 3D Gaussian primitives, turning them into 2D Gaussian primitives, which geometrically resemble discs. During the splatting process, 2DGS abandons the local affine transformation method used in 3DGS and introduces the ray-splat intersection method. This method directly computes the intersection position $\bm{u}(\bm{x})$ between the ray from the camera origin to the pixel and the 2D Gaussian primitives. The Gaussian value $G^{2D}$ is then calculated using a low-pass filter, specifically $\max \left\{ G\left(\mathbf{u}(\mathbf{x})\right), G\left(\frac{\mathbf{x} - \mathbf{c}}{\sigma}\right) \right\}$. The alpha blending formula is the same as in Eq.~\eqref{eq:rendering}, with $G^{3D}$ replaced by $G^{2D}$.

\subsection{Overview}
We propose a differentiable mixed 2D-3D Gaussian splatting method to reconstruct a realistically and geometrically accurate head avatar from multi-view 2D images, as shown in Fig.~\ref{fig:pipeline}. 
We begin by training 2D Gaussian representations of the 3D head avatar to ensure geometric accuracy. Next, we identify the 2D Gaussians for which the rendering quality of 2DGS is inadequate. For these identified 2D Gaussians, we bind additional 3D Gaussians to construct the mixed 2D-3D Gaussians in Sec.~\ref{sec:mixed}. We use a local-to-global transformation method to convert mixed 2D-3D Gaussians for creating dynamic 3D head avatars. Finally, we train the mixed 2D-3D Gaussians to create realistic and geometrically accurate head avatars. The progressive training strategy is introduced in Sec.~\ref{sec:training}. 

\subsection{Mixed 2D-3D Gaussian Avatar} \label{sec:mixed}
\subsubsection{Mixed 2D-3D Gaussian Modeling} \label{sec:modeling}
We introduce mixed 2D-3D Gaussian modeling, utilizing the 2D Gaussian to accurately represent geometric surfaces and the 3D Gaussian to enhance color representation. 
Specifically, we first train the model using 2DGS. 
Initially, we assign a 2D Gaussian to each triangular mesh of FLAME. As the adaptive density control process progresses, more generated 2D Gaussians will be bound to the mesh triangles.
After the 2DGS training, we take a rendered image $\hat{X}$ and a real image $X$ and divide each of them into tiles of size $w \times h$. Each tile is represented as $\hat{x}_{ij}$ for the rendered image and $x_{ij}$ for the real image, where $i$ and $j$ indicate the position indices of the tiles. The MSE loss value is defined as follows:
\begin{equation}
L_{ij} = \|\hat{x}_{ij} - x_{ij}\|_2^2
\end{equation}

\begin{figure}[t]
    \centering
    \includegraphics[width=0.5\textwidth]{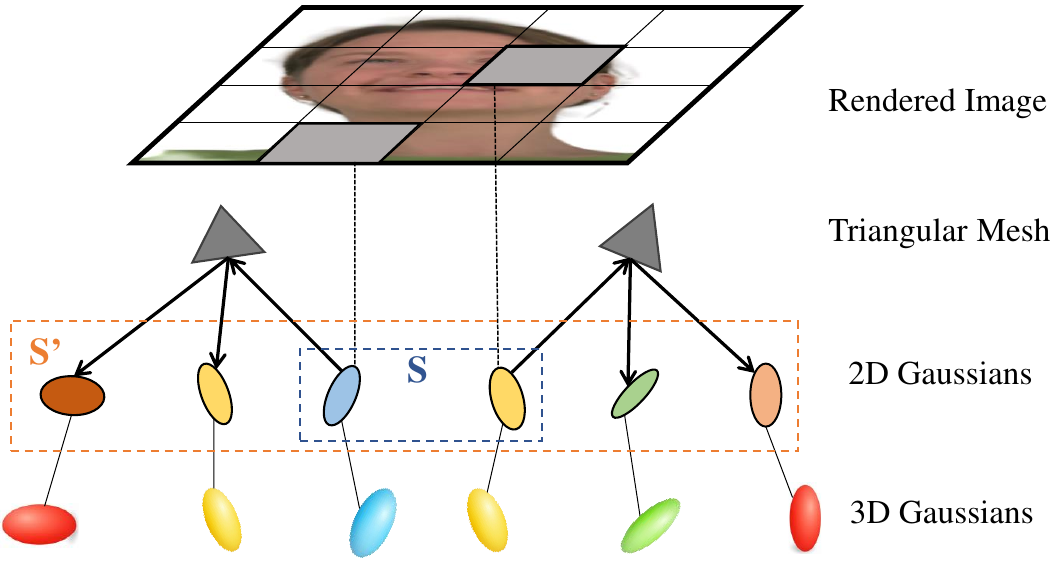}
    \caption{The tree structure of mixed 2D-3D Gaussians.}
    \label{fig:tree}
\end{figure}

We select the top $k$ tiles with the largest $L_{ij}$ values and record the indices of the 2D Gaussians contributing to the colors of these tiles, forming a set $S$. 
Since a triangle consists of multiple bounded 2D Gaussians, a Gaussian tree structure is formed with the triangular mesh as the root node and the 2D Gaussians as its child nodes, as shown in Fig.~\ref{fig:tree}. 
Based on the tree, for each 2D Gaussian in set \( S \), we add its sibling nodes to collectively form a larger set $S'$.
This process is repeated for $n$ iterations, with each iteration using different FLAME parameters and camera parameters to generate 2D images from multiple angles and with various expressions. The set $S'$ from each iteration is combined into a union set $U$. The whole pipeline is shown in Alg.~\ref{alg:selection}. 
\begin{equation}
U = \bigcup_{i=1}^{n} S'_i
\end{equation}

To improve the rendering quality of 2D Gaussians, we add the additional 3D Gaussians to the scene based on the set \( U \), as shown in Fig.~\ref{fig:tree}.
We treat each 2D Gaussian \( g^{2D} \in U \) as a parent node and generate a corresponding 3D Gaussian \( g^{3D} \) as its child node. These 3D Gaussians inherit the position $\mu$, spherical harmonics (SHs), opacity $\alpha$, rotation $r$, and scaling $s$  from corresponding parent \( g^{2D} \) in local space, with the third dimension of scaling initialized to an average value. In this way, we generate some 3D Gaussians in areas where the rendering quality of 2DGS is poor. No model parameter optimization is required at this stage.

\begin{algorithm}
\caption{Error-based 2D Gaussian Selection}
\label{alg:selection}
\begin{algorithmic}[1]
\Require Ground truth image $X$
\Require Number of tiles $w \times h$
\Require Number of iterations $n$
\Require Number of top tiles $k$
\Require Set of FLAME and camera parameters $f$, $c$
\State Initialize set $U \gets \emptyset$

\For{iteration $= 1$ to $n$}
    \State Sample $f$ and $c$
    \State Render image $\hat{X}$ using $f$ and $c$
    \State Divide images $\hat{X}$ and $X$ into $w \times h$ tiles
    \State Calculate MSE for each tile: $L_{ij} = \|\hat{x}_{ij} - x_{ij}\|_2^2$
    \State Select top $k$ tiles with highest $L_{ij}$ loss
    \State Record indices of 2D Gaussians contributing to these tiles in set $S$
    \State Based on the Gaussian tree, for each 2D Gaussian in set \( S \), add its sibling nodes to form set \( S' \)
    \State Update $U \gets U \cup S'$
\EndFor
\State \Return $U$
\end{algorithmic}
\vspace{-2pt}
\end{algorithm}
\subsubsection{Mixed 2D-3D Gaussian Splatting}
The splatting methods of 3DGS and 2DGS are entirely different; however, both utilize the same alpha-blending approach. By combining these two splatting methods, we project 3D Gaussians onto image space using a local affine transformation to obtain Gaussian values, while 2D Gaussians are computed directly using the ray-splat intersection method. 

During the rasterization process, for a given pixel \( \mathbf{x} \) belonging to tile \( t \), we first sort all the 2D Gaussians within tile \( t \) by depth and then perform alpha blending in order from front to back. Specifically, while traversing the 2D Gaussians, if the current 2D Gaussian \( g_{k}^{2D} \) has a 3D Gaussian \( g_{j}^{3D} \) child node, we insert the contribution of \( g_{j}^{3D} \) after that of \( g_{k}^{2D} \), in order to emphasize the compensation of the 3D Gaussians to the color. The mixed alpha blending formula is as follows:
\begin{equation}
T^{2D} = \prod_{t=1}^{i-1} \left(1 - \alpha_t G^{2D}_t(\mathbf{x})\right) 
\end{equation}
\begin{equation}
\mathbf{c}_{p}^{2D}(\mathbf{x}) = \sum_{i=1}^{k} \mathbf{c_i} \alpha_i G^{2D}_i(\mathbf{x}) T^{2D}
\end{equation}
\begin{equation}
\mathbf{c}^{3D}(\mathbf{x}) = \mathbf{c_j} \alpha_j G^{3D}_j(\mathbf{x}) \prod_{t=1}^{k} \left(1 - \alpha_t G^{2D}_t(\mathbf{x})\right) 
\end{equation}
\begin{equation}
\mathbf{c}_{b}^{2D}(\mathbf{x}) = \sum_{i=k+1}^{N} \mathbf{c_i} \alpha_i G^{2D}_i(\mathbf{x})  \left(1 - \alpha_j G^{3D}_j(\mathbf{x})\right)  T^{2D} 
\end{equation}
where \( G^{2D} \) represents the Gaussian value of 2D Gaussian, and \( G^{3D} \) represents that of 3D Gaussian. \( \mathbf{c}_{p}^{2D}(\mathbf{x}) \) denotes the color contribution to pixel \( \mathbf{x} \) from \( g_{k}^{2D} \) and all preceding 2D Gaussians, while \( \mathbf{c}_{b}^{2D}(\mathbf{x}) \) denotes the color contribution from the 2D Gaussians that follow \( g_{k}^{2D} \). \( \mathbf{c}^{3D}(\mathbf{x}) \) represents the color contribution from \( g_{j}^{3D} \). We add all color contribution together and obtain the final color of pixel \( \mathbf{x} \).
\begin{equation}
\mathbf{c}^{mixed}(\mathbf{x}) = \mathbf{c}_{p}^{2D}(\mathbf{x}) + \mathbf{c}^{3D}(\mathbf{x}) + \mathbf{c}_{b}^{2D}(\mathbf{x})
\end{equation}

\subsubsection{Local-to-Global Transformation}
To create dynamic 3D head avatars, we propose a novel local-to-global transformation method designed for mixed 2D-3D Gaussians, inspired by the 3D Gaussian rigging approach in GaussianAvatars~\cite{qian2024gaussianavatars}. The core of the local-to-global transformation is to establish a mapping between the FLAME mesh and the mixed Gaussians. As shown in the Gaussian tree relation in Fig.~\ref{fig:tree}, we associate each 2D Gaussian in local space with the triangular mesh in global space. Likewise, we connect each 3D Gaussian in local space to its corresponding 2D Gaussian.
The FLAME parameters, such as expressions, enable the canonical FLAME to transform into a deformable FLAME. Through this mapping, the triangular mesh can adjust along with the mixed Gaussians, allowing these mixed Gaussians to be represented in a global space.

In local space, we initialize the position $\mu_l$ of the mixed Gaussians to 0, the rotation matrix $r_l$ to the identity matrix, scaling $s_l$ to the unit vector, and opacity $\alpha$ to 0. 
In global space, we define the mixed Gaussians' position as $\mu_g$, rotation matrix as $r_g$, scaling as $s_g$, and opacity as $\alpha$.
For the transformation of mixed Gaussians through rotation and scaling, the formula is as follows:
\begin{equation}
r_g = Rr_l, s_g = \lambda s_l
\label{eq:r_tran}
\end{equation}
where $R$ is the rotation matrix formed by the direction vector of one edge of the triangle, the normal vector, and their cross-product. $\lambda$ represents the average length of an edge of the triangle and its perpendicular, serving as a global position and size scaling factor.

For the mapping between the mesh and the 2D Gaussian, the transformation of positions is defined as follows:
\begin{equation}
\mu_g^{2D} = \lambda R \mu_l^{2D} + T + p^{2D}_\theta(t)
\end{equation}
where $T$ denotes the average position of the three vertices of the corresponding triangle in the global space. We introduced a learnable perturbation term \( p^{2D}_\theta(t) \) to adaptively correct positional errors, where $t$ represents the timestep.

For the mapping between 2D and 3D Gaussians, we substituted $T$ with $\mu^{2D}$ in the position transformation formula:
\begin{equation}
\mu_g^{3D} = \lambda R \mu_l^{3D} + \mu_g^{2D} + p^{3D}_\theta(t)
\end{equation}

\begin{figure*}[!ht]
    \centering
    \includegraphics[width=0.95\textwidth]{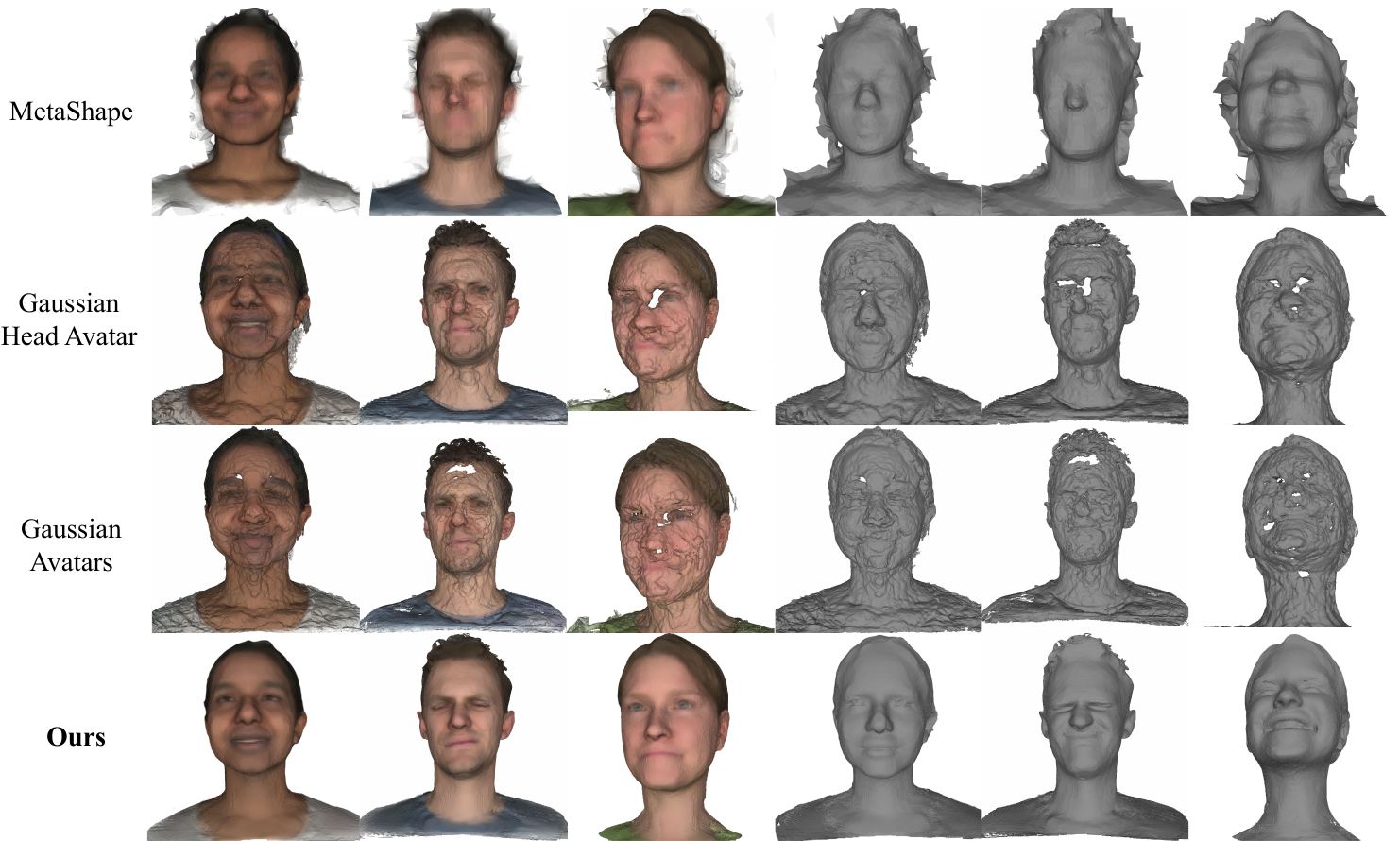}
    \vspace{-2mm}
    \caption{Qualitative comparison on mesh reconstruction.}
    \label{fig:vis_recon}
    \vspace{-2mm}
\end{figure*}

\subsection{Progressive Training Strategy} \label{sec:training}
\subsubsection{2D Gaussian Training} \label{sec:2d_training}
At this stage, we will only use 2DGS to train the model, with the goal of achieving a highly accurate geometric surface. The loss functions in this stage align with those of GaussianAvatars and 2DGS. We calculate the $L_1$ loss between the rendered images and the ground truth, including a D-SSIM term. In this case, $\lambda$ is set to 0.2.
\begin{equation}
L_{\text{rgb}} = (1 - \lambda)L_1 + \lambda L_{\text{D-SSIM}}
\end{equation}

To ensure that the locations of the 2D Gaussians and the corresponding underlying mesh remain adequately aligned in global space, we constrain the local position $\mu_l$ using $L_{\text{pos}}$. The threshold $\epsilon_{\text{pos}}$ is 1.
\begin{equation}
L_{\text{pos}} = \left\| \max \left( \mu_l^{2D}, \epsilon_{\text{pos}} \right) \right\|_2
\end{equation}

Likewise, the scaling of the 2D Gaussians should correspond to that of the triangular mesh. To prevent the local scaling $s_l$ from becoming excessively large, we constrain it using $L_{\text{sca}}$, with $\epsilon_{\text{sca}}$ set to 0.6.
\begin{equation}
L_{\text{sca}} = \left\| \max \left( s_l^{2D}, \epsilon_{\text{sca}} \right) \right\|_2
\end{equation}

To improve the performance of 2DGS in geometric modeling, we utilize their depth distortion function $L_{\text{d}}$ and normal consistency function $L_{\text{n}}$ (details can be found in ~\cite{huang20242d}). Therefore, the training loss in this stage is expressed as:
\begin{equation}
L_{\text{rec}} = L_{\text{c}} + \lambda_1 L_{\text{pos}} + \lambda_2 L_{\text{sca}} + \lambda_3 L_{\text{d}} + \lambda_4 L_{\text{n}}
\end{equation}

\subsubsection{Mixed 2D-3D Gaussian Training}
After completing training for 2DGS in Sec.~\ref{sec:2d_training} and mixed 2D-3D Gaussian modeling in Sec.~\ref{sec:modeling}, the scene now includes 2D Gaussians distributed across the surface of the geometry and 3D Gaussians created for color compensation. During the mixed 2D-3D Gaussian training stage, we train all parameters of the 3D Gaussians as well as the color parameters (spherical harmonics) of the 2D Gaussians. However, mixed 2D-3D Gaussians require that each 3D Gaussian is positioned close to its corresponding parent 2D Gaussian in global space. This proximity ensures that when both are projected into image space, they can significantly contribute to the same tile's color. We propose a 2D-3D distance regularization term \( L_{dis} \) that limits the distance between 2D and 3D Gaussians to a specified threshold. \( \epsilon_{dis} \) is 1.
\begin{equation}
L_{\text{dis}} = \left\| \max \left( \mu_l^{3D}+p_{\theta}^{3D}, \epsilon_{\text{dis}} \right) \right\|_2
\end{equation}
Finally, the training loss in this stage is expressed as:
\begin{equation}
L_{mixed} = L_{rgb} + \lambda_5 L_{dis}
\end{equation}
where $\lambda_1$ = $\lambda_5$ = 0.01, $\lambda_2$ = 1, $\lambda_3$ = 1000, and $\lambda_4$ = 0.05.

\section{Experiments}

\begin{figure*}[!ht]
    \centering
    \includegraphics[width=0.9\textwidth]{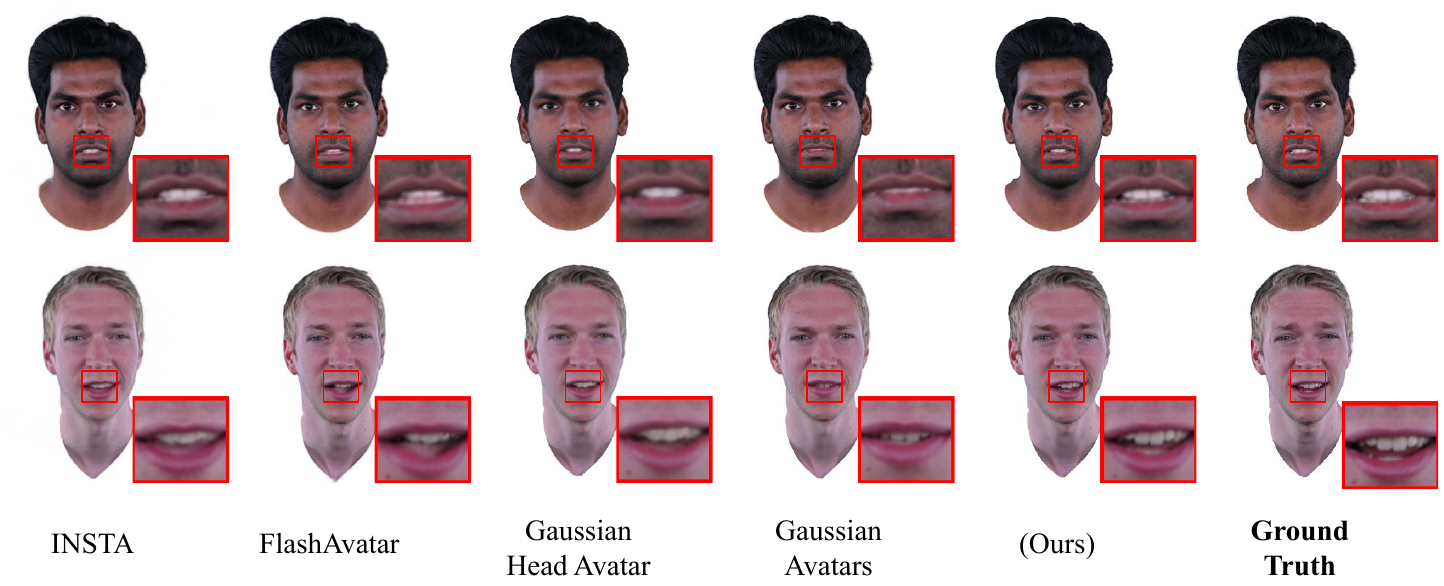}
    \vspace{-2mm}
    \caption{Qualitative comparison of image rendering. }
    \label{fig:vis_rendering}
    \vspace{-2mm}
\end{figure*}

\subsection{Experimental Settings}
\subsubsection{Dataset and Baselines} 
We use two challenging datasets: NeRSemble and INSTA.
NeRSemble dataset~\cite{kirs2023nersemble} is used to conduct reconstruction and rendering experiments on 9 subjects, with each recording including 16 viewpoints. 
To ensure a fair comparison, we use the same dataset split as GaussianAvatars. 
INSTA dataset~\cite{zielonka2023instant} is used for rendering experiments. It consists of monocular videos from ten subjects, with the final 350 frames of each sequence set aside for testing.

For the NerSemble dataset, we selected NeRF-based and its variant methods, including PointAvatar~\cite{zheng2023pointavatar} and INSTA~\cite{zielonka2023instant}; and 3DGS-based methods, including Gaussian Head Avatar~\cite{xu2024gaussian} and GaussianAvatars~\cite{qian2024gaussianavatars}.
For the INSTA dataset, we selected NeRF-based methods, including NHA, IMAvatar~\cite{zheng2022avatar}, and INSTA; and 3DGS-based methods, including FlashAvatar, Gaussian Head Avatar, and GaussianAvatars.
Additionally, we selected a 3D reconstruction industrial software, MetaShape~\cite{agisoft2023} as a baseline.

\subsubsection{Evaluation Metrics}
Due to the absence of ground truth for 3D head meshes in the datasets, we are unable to utilize quantitative metrics to evaluate geometric reconstruction accuracy. Our method demonstrates significant advantages in the visual quality of geometric reconstruction compared to the baselines; therefore, we will evaluate it using visual comparison.

In terms of rendering, to assess the quality of the synthesized images, we utilized common metrics such as Mean Squared Error (L2), Peak Signal-to-Noise Ratio (PSNR), Structural Similarity Index (SSIM), and Learned Perceptual Image Patch Similarity (LPIPS).

\subsubsection{Mesh Extraction and Texture Generation}
To extract meshes from the reconstructed mixed Gaussians, we generate depth maps of the training views using the median depth values of the 2D Gaussians projected onto the pixels. 
To achieve a high-quality texture for the 3D head mesh, we utilize the RGB color from the mixed 2D-3D Gaussians projected onto the pixels. 
Then, we use Open3D~\cite{zhou2018open3d} to apply truncated signed distance fusion (TSDF) to merge the depth maps for mesh extraction and use the RGB color to generate the texture. For a fair comparison, we apply the same mesh extraction and texture generation method to the 3DGS-based baselines.

\subsubsection{CUDA Implementation}
\begin{figure}[h]
    \centering
    \vspace{-6mm}
    \includegraphics[width=0.5\textwidth]{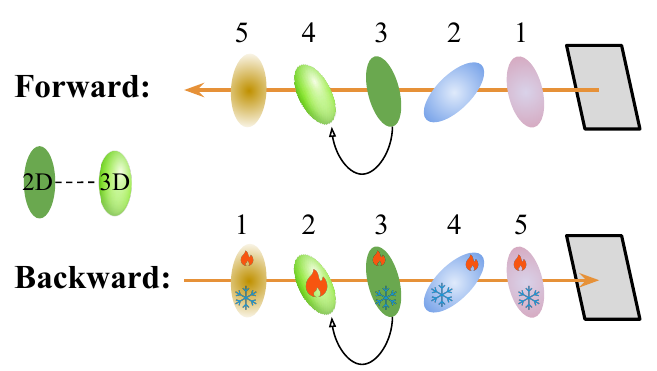}
    \vspace{-6mm}
    \caption{The CUDA implementation of mixed 2D-3D Gaussian Splatting.}
    \label{fig:for_back}
\end{figure}

\begin{figure}[h]
    \centering
    \vspace{-6mm}
    \includegraphics[width=0.3\textwidth]{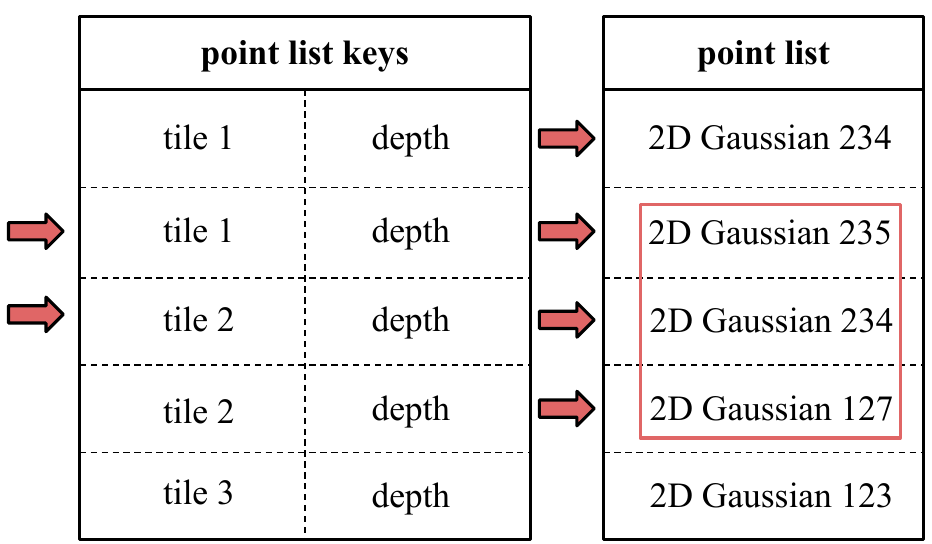}
    \vspace{-2mm}
    \caption{The CUDA implementation of error-based selection.}
    \label{fig:point_list}
\end{figure}

We have implemented mixed 2D-3D Gaussian Splatting with CUDA kernels based on the 2DGS~\cite{huang20242d} and 3DGS~\cite{kerbl20233d} frameworks. 

In the implementation of error-based selection in mixed 2D-3D Gaussian modeling, we extract the point list and point list keys from the 2DGS CUDA rasterizer. These two lists are one-to-one correspondences, with each element corresponding to the projection of a Gaussian in the image space, and they are ordered by depth. Each element in the point list keys is a 64-bit integer, where the first 32 bits store the tile's ID, and the last 32 bits store the depth value of the current Gaussian projection. Each element in the point list is a 32-bit integer, which stores the index of the 2D Gaussian corresponding to the Gaussian projection. Therefore, using these two lists, we can establish the mapping between the tiles and the 2D Gaussians. By using the tile's ID, we can identify the 2D Gaussians that contribute to it, as shown in line 8 of Alg.~\ref{alg:selection}. 
For example, as shown in Fig.~\ref{fig:point_list}, after the 2D Gaussian training process, we calculate that tiles 1 and 2 have the largest MSE. Based on the point list keys and point list, we have identified the corresponding 2D Gaussian indices as 235, 234, and 127.

In the implementation of mixed 2D-3D Gaussian splatting, we pass the connection between 2D and 3D Gaussians to the CUDA rasterizer. Fig.~\ref{fig:for_back} is a schematic diagram of this process.
During the forward pass, for each pixel, we perform alpha blending of the projections of all 2D Gaussians in the current tile in depth order, from front to back. When a 2D Gaussian with a 3D Gaussian child node is encountered, the projection of the 3D Gaussian is inserted after the projection of the current 2D Gaussian, and alpha blending continues.
In the backward pass, this order is reversed, meaning that alpha blending is performed from back to front. Therefore, the projection of the child 3D Gaussian will be inserted in front of its parent 2D Gaussian projection, and the alpha blending of the child 3D Gaussian will be computed first. During the backward pass, we update the gradients of all parameters for the 3D Gaussians, as well as the gradients for the color parameters of the 2D Gaussians.

\subsection{Comparison Results}
Comparison experiments are divided into quantitative and qualitative experiments. 
In our quantitative experiments, we assessed the quality of image rendering using two different datasets.
In the qualitative experiments for reconstruction, we presented two visualization results: mesh without texture and mesh with texture. For rendering, we demonstrate the results of the generated RGB images.

\begin{table*}[!ht]
    \centering
    \caption{Results on the NeRSemble~\cite{kirs2023nersemble} and INSTA~\cite{Zielonka2022InstantVH} dataset. Green indicates the best and yellow indicates the second.}
    \resizebox{0.70\textwidth}{!}{%
    \begin{tabular}{l|c|cccc} 
    \toprule
         Method & dataset & L2 $\downarrow$ & PSNR $\uparrow$ & SSIM $\uparrow$ &  LPIPS $\downarrow$  \\
         \midrule
         PointAvatar~\cite{zheng2023pointavatar}& \multirow{5}{*}{NeRSemble} & 0.0020 & 25.9 & 0.887 & 0.096  \\
         INSTA~\cite{Zielonka2022InstantVH} & & 0.0018 & 26.8 & 0.895 & 0.117   \\
         Gaussian Head Avatar~\cite{xu2024gaussian} & & 0.0015 & 30.8 & \cellcolor{yellow!30} 0.945 & 0.078  \\
         GaussianAvatars~\cite{qian2024gaussianavatars} & & \cellcolor{yellow!30} 0.0013 & \cellcolor{yellow!30} 31.4 &  0.941 & \cellcolor{green!30}0.064  \\
         (Ours) & & \cellcolor{green!30} 0.0011 & \cellcolor{green!30} 31.8 & \cellcolor{green!30} 0.953 & \cellcolor{yellow!30} 0.067  \\
         \midrule
         NHA~\cite{grassal2022neural} & \multirow{7}{*}{INSTA} & 0.0024 & 27.0 & 0.942 & 0.043  \\
         IMAvatar~\cite{zheng2022avatar} & & 0.0021 & 27.9 & 0.943 & 0.061  \\
         INSTA~\cite{Zielonka2022InstantVH} &  & 0.0017 & 28.6 & 0.944 & 0.047  \\
         FlashAvatar~\cite{xiang2024flashavatar} &  & 0.0017 & 29.2 & 0.949 & \cellcolor{yellow!30}0.040  \\
         Gaussian Head Avatar~\cite{xu2024gaussian} &  & 0.0016 & \cellcolor{yellow!30}29.7 & \cellcolor{yellow!30}0.958 & 0.043  \\
         GaussianAvatars~\cite{qian2024gaussianavatars} &  & \cellcolor{yellow!30}0.0014 & 29.1 & 0.953 & 0.046  \\
         (Ours)  & & \cellcolor{green!30}0.0012 & \cellcolor{green!30}30.4 & \cellcolor{green!30}0.962 & \cellcolor{green!30}0.039   \\
         \midrule

    \end{tabular}
    }
    \label{tab:main_result}
\end{table*}

\subsubsection{Quantitative Evaluation}
In this experiment, NeRSemble was used to assess novel-view synthesis, while INSTA dataset was used for evaluating self-reenactment.
As shown in Tab.~\ref{tab:main_result}, it is clear that NeRF-based and its variant methods, such as PointAvatar and INSTA, generally yield lower image quality. In contrast, methods based on 3DGS tend to demonstrate stronger rendering capabilities. This observation highlights the inherent advantage of 3D Gaussians in color representation.
FlashAvatar binds 3D Gaussians to the UV map, and controlling the number of Gaussians can lead to insufficient detail representation, resulting in relatively lower rendering quality compared to other 3DGS-based methods.
``Ours w/o 3DGS'' in Tab.~\ref{tab:ablation} presents the rendering results of our method on the NeRSemble dataset utilizing only 2D Gaussians. It is evident that, in comparison to methods based on 3DGS, the multi-view consistency of 2DGS greatly diminishes image quality.
However, our method employs 3D Gaussians to overcome the limitations of 2D Gaussians in color rendering, resulting in higher-quality rendering than pure 2DGS. Our approach is highly competitive with 3DGS methods, but it significantly excels in geometric surface accuracy, outpacing 3DGS-based methods.

\subsubsection{Qualitative Evaluation}

For accurate mesh reconstruction, we assess visual results using the multi-view NeRSemble dataset, as our datasets lack ground truth for reconstructed meshes. We presented the results of visualizing 3D head avatar reconstruction using our method alongside representative baselines. As shown in Fig.~\ref{fig:vis_recon}, our reconstruction results emphasize the mesh and texture, depicted as mesh without and with texture, respectively.
In the mesh w/o texture visualization, we mainly evaluate the accuracy of the geometric surface reconstruction.
Methods based on 3DGS exhibit the poorest reconstruction quality, resulting in uneven mesh surfaces and significant noise. This is due to the inherent multi-view inconsistency of 3DGS and the contradiction between its three-dimensional shape and the two-dimensional mesh surface. 
The face reconstructed by MetaShape is blurry, and it produces many irrelevant floating artifacts.
Our approach creates detailed and noise-free surfaces by utilizing 2D Gaussians to preserve the 3D head surface.
In the mesh w/ texture visualization, MetaShape and methods based on 3DGS struggle with texture color accuracy due to significant errors in the geometric surface reconstruction.

Fig.~\ref{fig:vis_rendering} offers a qualitative comparison of the latest baselines for more accurate rendering. We have highlighted the comparative details using red boxes. Our method significantly outperforms INSTA in terms of facial detail representation. Mixed 2D-3D Gaussians show strong competitiveness compared to methods based on 3DGS, and in some cases, they even outperform these methods.

\begin{table}[!t]
    \centering
    \caption{Quantitative ablation study on NeRSemble dataset.}
    \resizebox{1.0\linewidth}{!}{%
    \begin{tabular}{l|c@{\hskip 0.1in}c@{\hskip 0.1in}c@{\hskip 0.1in}c@{\hskip 0.1in}}
    \toprule
      Method   & L2 $\downarrow$  & PSNR $\uparrow$ & SSIM $\uparrow$ & LPIPS $\downarrow$ \\
      \midrule
    Ours \small{w/o 3DGS} & 0.0016 & 27.8  & 0.943  & 0.079 \\
    Ours \small{w/o $p_{\theta}$} & 0.0013 & 30.9 & 0.945  & 0.072 \\
    Ours \small{w/o $L_{dis}$}  & 0.0013 & 29.7 & 0.946  & 0.078 \\
    Our full pipeline & \textbf{0.0011}  & \textbf{31.8} & \textbf{0.953} & \textbf{0.067} \\
     \bottomrule
    \end{tabular}
    }
    \label{tab:ablation}
\end{table}

\begin{figure}[!t]
    \centering
    \includegraphics[width=1.0\linewidth]{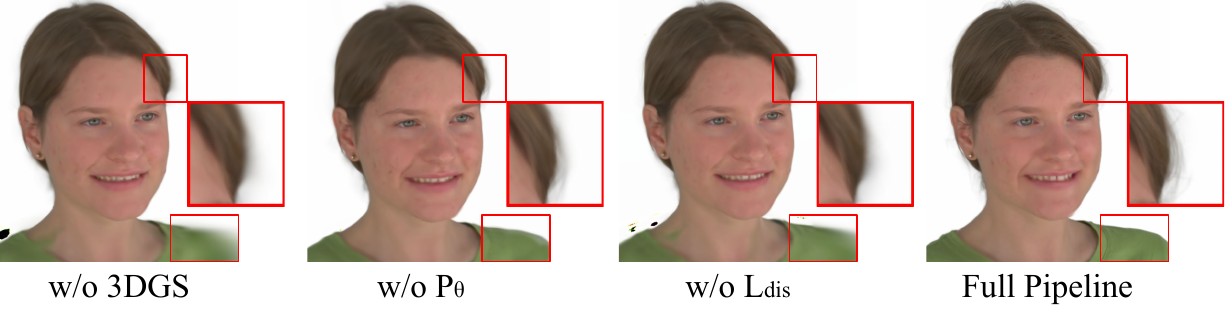}
    \caption{Qualitative ablation study on the NeRSemble dataset. 
    }
    \label{fig:ablation}
    \vspace{-5mm}
\end{figure}

\subsection{Ablation Study}
In this section, we conducted an ablation study by component to evaluate the role of each element on the NeRSemble dataset. We trained the model using only 2DGS for all iterations without using 3DGS, referred to as ``w/o 3DGS," to evaluate the contribution of 3DGS to RGB rendering. 
We removed the perturbation term $p_{\theta}$ in local-to-global transformation, denoted as ``w/o $p_{\theta}$." 
Additionally, we removed the 2D-3D distance loss $L_{dis}$, denoted as ``w/o $L_{dis}$".

The results are shown in Tab.~\ref{tab:ablation} and Fig.~\ref{fig:ablation}. the absence of all components results in a decrease in rendering quality, indicating the effectiveness of these components.
Removing 3D Gaussians resulted in a significant drop in rendering quality, demonstrating that mixed Gaussians have a considerable effect in maintaining high-precision color rendering. 
The learnable \( p_{\theta} \) enables a trade-off between 2D Gaussians moving towards or away from the triangular mesh, allowing for adaptive correction of spatial positioning errors. The loss \( L_{dis} \) maintains the distance between 2D Gaussians and their corresponding 3D Gaussians in global space, ensuring that the projection of the 3D Gaussians after local affine transformation lies within the same tile as the 2D splats, thereby compensating for color errors to the greatest extent.

\section{Conclusion}
In this work, we introduce MixedGaussianAvatar, a new method that employs mixed 2D-3D Gaussian Splatting for creating head avatars. We use 2DGS to maintain the surface geometry and employ 3DGS for color correction in areas where the rendering quality of 2DGS is insufficient, reconstructing a realistically and geometrically accurate 3D head avatar. 
We employ a local-to-global transformation technique to reconstruct dynamic head avatars and adopt a progressive training strategy to train mixed 2D-3D Gaussians. 
Our method unifies image rendering and mesh reconstruction, using a mixed Gaussian representation to integrate the two modalities of 2D images and 3D meshes. It bridges multimedia expressions and multimodal interactions, seamlessly unifying 2D visual fidelity with 3D geometric authenticity for holistic avatar representation.

\bibliographystyle{ACM-Reference-Format}
\balance
\bibliography{sample-base}


\end{document}